# Numerical Weather Prediction (NWP) and hybrid ARMA/ANN model to predict global radiation

Cyril Voyant[1,2*], Marc Muselli[1], Christophe Paoli[1], Marie-Laure Nivet[1]

[1] University of Corsica, CNRS UMR SPE 6134, 20250 Corte, France
[2] Castelluccio Hospital, Radiotherapy Unit, BP 85, 20177 Ajaccio, France

*Corresponding author: Cyril Voyant, tél: +33 4 95 29 36 66, fax:+ 33 4 95 29 37 97, email: cyrilvoyant@gmail.com

**Abstract**.

We propose in this paper an original technique to predict global radiation using a hybrid ARMA/ANN model and data issued from a numerical weather prediction model (ALADIN). We particularly look at the Multi-Layer Perceptron. After optimizing our architecture with ALADIN and endogenous data previously made stationary and using an innovative pre-input layer selection method, we combined it to an ARMA model from a rule based on the analysis of hourly data series. This model has been used to forecast the hourly global radiation for five places in Mediterranean area. Our technique outperforms classical models for all the places. The nRMSE for our hybrid model ANN/ARMA is 14.9% compared to 26.2% for the naïve persistence predictor. Note that in the stand alone ANN case the nRMSE is 18.4%. Finally, in order to discuss the reliability of the forecaster outputs, a complementary study concerning the confidence interval of each prediction is proposed

# NOMENCLATURE

| | *Time series generalities* | | *ANN nomenclature* |
|---|---|---|---|
| $x_t / \hat{x}_t$ | Time series measure *and prediction at time t* | $B^1$ | $b_i^1$ set, values of the biases of the hidden nodes |
| $X_t / \hat{X}_t$ | Global time series measure *and prediction at time t (Wh/m²)* | $B^2$ | $b_i^2$ set, value of the output node bias |
| $\omega_1/\omega_2$ | Angular frequencies of a signal (s$^{-1}$) | $H$ | Number of hidden nodes |
| $R_t$ | Time series calculated from global radiation and representing the cloud cover occurrence | $In$ | Number of endogenous input nodes |
| $a(t)/b(t)$ | Continuous functions representing the amplitude of the global radiation *(Wh/m²)* | $f/g$ | Activation function of the output and hidden layer |
| | | $\omega_{ij}^1$ | values of the weight between input nodes *j* and hidden nodes *i* |
| $f_n$ | Regression model group | $\omega_i^2$ | weight values between hidden node *i* and output node |
| $\epsilon_{t+1}$ | Residue of regression at time *t+1 t* | $\Delta\omega$ | wieghts optimization for the Levenberg-Marquard algorithm |
| $E[X(t)]$ | Expected value of *X(t)*; the first moment of *X(t)* is the average, and the second the variance | $J$ | Jacobian matrix |
| $\gamma(h)$ | Covariance between two variables without time dependence | $I$ | Identity matrix |
| | *Time series preprocessing* | $e(\omega)$ | Error prediction vector |
| $H_0'(t)$ | Corrected extraterrestrial solar radiation coefficient at time t [Wh/m²] | $\mu$ | Specific algorithm parameter |
| $H_{gh}$ | Clear sky global horizontal radiation [Wh /m²] | $endo^e$ | e lags endogenous for ANN |
| $\tau$ | Global total atmospheric optical depth | $\hat{N}(t), n$ | ALADIN forecast of the nebulosity at time *t* and *n*, the number of nebulosity lag in ANN input [octas] |
| $b$ | Fitting parameter of the Solis clear sky model | $\hat{P}(t), p$ | ALADIN forecast of the pressure at time *t* and *ps*, the number of pressure lag in ANN input [Pa] |
| $h$ | Solar elevation angle at time *t* (rad) | $\hat{T}(t), t$ | ALADIN forecast of the temperature at time *t* and *tp*, the number of temperature lag in ANN input [°C] |
| $CSI_t$ | Clear sky index at time *t* | $\widehat{RP}(t), rp$ | ALADIN forecast of the rain precipitation at time *t* and *rp*, the number of precipitation lag in ANN input [mm] |
| $CSI_t^*$ | Clear sky index with seasonal adjustments at time *t* and related to *i* (period number; number of year, or days) and *j* (number of measures per period). | $H$ | Number of hidden nodes in the MLP |
| $CSI^*_{i,j}$ | | $In$ | Number of input nodes in the MLP |
| $MM_t$ | Moving average of *CSI* at time *t* | | *Linear regression* |
| $C_t$ | Periodic coefficients at time *t* | $W^{LR}$ | $w_j^{LR}$ set representing a vector of the linear regression coefficients |
| | *Statistical tests* | $Y$ | Vector with the *CSI*(t+1)* values |
| $VC_X$ | Variation coefficient (%) | $S'$ | Matrix with concatenated endogenous and exogenous data |
| $F_c$ | Fisher statistic | | ***ARMA model*** |
| $V_p$, $p$ | Variance of the period, and $p$ number of measures per period | $p$, $q$ | Order of ARMA(p,q) |
| $V_R$, $N$ | Variance of the residue and $N$ the number of period | $L$ | Lag operator |
| NbH | Number of hours of prediction contained during one year (365x9=3285) | $\varphi_i$, $\theta_i$ | $\varphi_t$ are the parameters of the AR model and $\theta_t$ the parameters of the MA model |
| CI(t), CI*(t) | Confidence interval for the time t of prediction. The star is related to the yearly average | | |

# 1. Introduction

Solar radiation is one of the principal energy sources, occupying a very important role in some engineering applications as production of electricity, heat and cold [1-4]. The process of converting sunlight to electricity without combustion creates power without pollution. It is certainly one of the most interesting themes in solar energy area. To use ideally this technology, it is necessary to understand and create efficient prediction models like done for example in Mueller et al, 2004, Mellit et al, 2005 and Mubiru et al, 2008 [5-7]. Insolation is defined as the solar radiation striking a surface at a certain time and place and is typically expressed in kilowatt hours per meter square (kWh/m²) [8,9]. Many factors determine how much sunlight is available at a given location. We can mention the atmospheric conditions, the Earth's position in relation to the sun, and the site obstructions [4,9]. Atmospheric conditions that can also affect the amount of radiation received on the Earth's surface are the quantity of air molecules, water vapor, dust, ozone and carbon dioxide, the cloud cover, the air pollution, the dust storms, the volcanic eruptions, etc. [5,10]. There is an interest to control the solar radiation prediction, as for example to identify the most optimal locations for developing solar power project or to maintain the grid stability in solar and conventional power management. Note that, in Europe, the White Paper on Energy (established in 1997) set a target (not yet achieved) of 12% of electricity production from renewable energies by 2010. The new European guidelines (in press) set a new target to 20% by 2020. The issues of the solar energy prediction are very important and mobilize a lot of research teams around the world and particularly in the Mediterranean area [3,7,11-13]. In practice, the global radiation (or insolation) forecasting is the name given to the process used to predict the amount of solar energy available in the current and near terms. A lot of methods have been developed by experts around the world [6,14,15]. Often the Times Series (TS) mathematical formalism is necessary. It is described by sets of numbers that measures the status of some activity over time [16]. In primary studies [1,13,17] we have demonstrated that an optimized multi-layer perceptron (MLP) with endogenous input made stationary and exogenous inputs (meteorological data) can forecast the global solar radiation time series with acceptable errors. This prediction model has been compared to other prediction methods (ARMA, k-NN, Markov Chains, etc.) and we have concluded that ANN and ARMA were similar. Following these studies and in order to see if we can significantly improve our results, we decided to add weather forecast (instead of exogenous data previously used) as new inputs of our mode. We assumed that Numerical Weather Prediction (NWP) simulations tools compute data patterns essential for determining solar radiation. These weather forecast data present two advantages: firstly they are becoming more and more available through the internet and secondly the models provide spatially distributed data which are very relevant to the regional scale studies. Thanks to an agreement with Météo-France, which is the French meteorological organization (http://france.meteofrance.com), we had the opportunity to freely access to some of the forecasts of the French operational limited area model ALADIN [18-20]. ALADIN (http://www.cnrm.meteo.fr/aladin) is a hydrostatic model developed by Météo-France in collaboration with the European Centre for Medium Range Weather Forecasts (ECMWF). It is the result of a project launched in 1990 by Météo-France with the aim of developing a limited area model and today 15 countries are participating in the common work. We propose in this paper an original technique to predict hourly global radiation time series using meteorological forecasts from the ALADIN NWP model. After optimizing our MLP with ALADIN forecast data and endogenous data previously made stationary with an ad-hoc method, we combine it to an Auto-Regressive and Moving Average (ARMA) model from rules based on the analysis of hourly data series. Finally we present all forecasting results with confidence intervals in order to give more complete information to a final user, such as a power manager.

The paper is organized as follow. Section 2 describes the data we have used: radiation time series measured from meteorological stations and forecast data computed by the ALADIN numerical weather model. After recalling the principles of time series forecasting and the need to make stationary a time series, we present in section 3 the forecasting models (ARMA and ANN) that have allowed us to build our original hybrid method, and the variables selection approach used. The section 4 includes final results and experiences conducted during this study and showing that forecasting results can significantly be improved by selecting ANN or ARMA models according to their performances. The reliability of the predictions is also considered by computing the confidence intervals. The section 5 concludes and suggests perspectives.

## 2. Radiation time series and Numerical Weather Prediction forecast data

In this study we have used two types of data: radiation time series and meteorological forecasts from the ALADIN NWP model. In order to verify the robustness of our approach we chose to apply our methodology on five distinct stations located in Mediterranean coastal area. Figure 1 shows the location in Mediterranean area of the five weather stations studied.

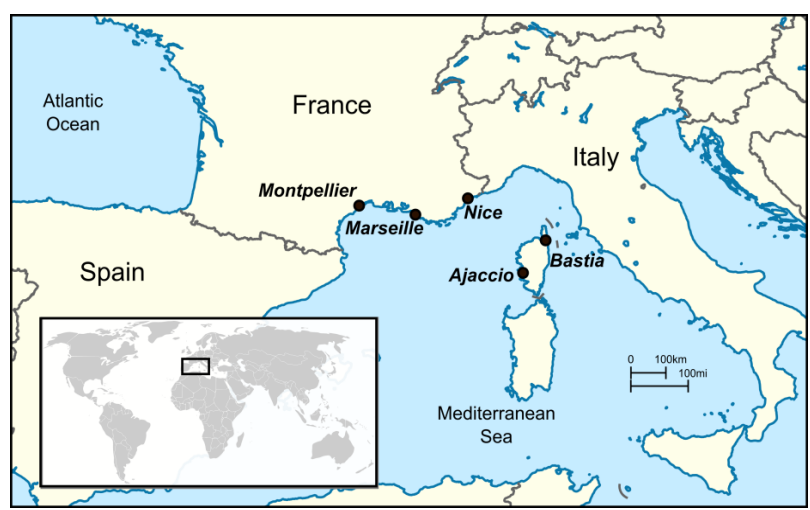

**Figure 1. The five studied stations marked in the Mediterranean sea: Ajaccio, Bastia, Montpellier, Marseille and Nice.**

Concerning endogenous data, the radiation time series (Wh.m$^{-2}$) are measured at coastal meteorological stations maintained by the French meteorological organization: Météo-France. We selected Ajaccio (41°5'N and 8°5'E, seaside, 4 m asl), Bastia (42°3'N, 9°3'E , 10 m asl), Montpellier (43.6°N and 3.9°E, 2 m asl), Marseille (43.4°N and 5.2°E, 5 m asl) and Nice (43.6°N and 7.2°E, 2 m asl). These stations are equipped with pyranometers (CM 11 from Kipp & Zonen) and standard meteorological sensors (pressure, nebulosity, etc.). The choice of these particular places is explained by their closed geographical and orographical configurations. All stations are located near the Mediterranean Sea with and mountains. This specific geographical configuration makes nebulosity difficult to forecast. Mediterranean climate is characterized by hot summers with abundant sunshine and mild, dry and clear winters. The data representing the global horizontal solar radiation were measured on an hourly basis from October 2002 to December 2008 (more than 6 years). The first four years have been used to setup our models and the last two years to test them. Note that in the presented study, only the hours between 8:00AM and 04:00PM (true solar time) are considered. The others hours are not interesting from energetically point of view and their predictions are complicated because it is very difficult to make stationary the measures of sunrise and sunset (stationarity scheme detailed in the next section). A first treatment allows us to clean the series of non-typical points related to sensor maintenances or absence of measurement. Less than 4 % of measurements were missing and replaced by the hourly average for the given hour.

The second types of data we decided to use are the meteorological forecasts from the ALADIN NWP model. Météo-France proposed us a free access to some of the forecasts issued from their numerical weather prediction model called ALADIN-France. This model is a bi-spectral limited area, based on the assimilation of daily measurements, and driven using, for boundary data, the outputs of the ARPEGE (acronym of "Action de Recherche Petite Echelle Grande Echelle") global model providing also by French meteorological services. The model evolves in average every six months because the ALADIN code follows the ARPEGE one in its permanent evolution and we are actually on the 37$^{th}$ cycle. For a better description of the model and its parameterization, the interested reader may refer e.g. to [18-20]. The French NWP system is organized around the production of analyses at 00, 06, 12 and 18UTC, and the range of the forecast is 54 hours. The horizontal resolution of ALADIN-France is approximately 9.5 Km, with 60 levels vertically. These data are those used actually, that is to say in the $37^{th}$ cycles. The ALADIN model has more than twenty outputs available at a high temporal resolution of one hour but not the global radiation. These values are computed in all points of the computing grid with mesh size of 9.5 Km. Considering these facts we had three major choices to do. First we had to select the forecast parameters to add as an input of our model. In a second time we had to choose the grid points for our five locations, and finally we add to select the analyses (between 00, 06, 12 and 18UTC) and ranges (1 to 54h) of the forecasts to take into account. For the ALADIN output, we based our choice on preceding works [17] in which we analyzed the benefit of taken into account exogenous variables. In these studies we made some computations about the correlation between the global radiation and a lot of exogenous meteorological parameters. Among the 23 ALADIN possible outputs we chose those which seem to have a straight link to solar radiation [17]. These data are pressure ($P$, Pa), nebulosity ($N$, Octas), rain precipitations ($RP$, mm) and temperature ($T$, °C). The second choice was on the selection of the grid points. We decided to simply use a proximity criterion and we chose the points of the ALADIN computational grid nearer to our five selected stations. We decided to take into account the analyses closer to the sunrise: we select the 06AM analyze with a 12 hours head forecast horizon. This couple of analyze and range allows us to cover, with a unique ALADIN extraction per day, the central hours we consider, that is to say 8:00AM to 4:00PM. However, we are aware that some of our choices can be discussed: first the periodic changes in the ALDIN model (more of ten cycles) may affect the training capability of the ANN, secondly an explicit help of a professional forecaster may comfort us in the choice of NWP model parameters, we will return to these restrictions in our conclusion.

## 3. Forecasting models

This section details the prerequisites that have enabled us to build our prediction model. After recalling the principles of time series forecasting and the need to make stationary a time series and how to validate it, we present the ARMA and ANN models. To conclude this section we present our final prediction model as a combination of ARMA and ANN models.

### a. Time series analysis, need of stationarity and validation

There are different approaches to model Time Series (TS) [16]: ARMA [15,21], ANN [22,23], are the ones we studied and which are the most effective based on our previous works. However the common base of all these models seems to be that a TS $x_t$ can be defined by a linear or non-linear model called $f_n$ (see Equation 1 where $t = n, n-1, ..., p+1, p$ with $n$, the number of observations and $p$ the number of parameters ; $n \gg p$) [24,28].

$$x_{t+1} = f_n(x_t, x_{t-1}, ..., x_{t-p+1}) + \epsilon_{t+1} \qquad (Eq\ 1)$$

To estimate the $f_n$ model, a stationarity hypothesis is often necessary. This condition usually implies a stable process [23-26]. This notion is directly linked to the fact that whether certain feature such as mean or variance change over time or remain constant. In fact, the time series is called weak-sense or weakly stationary if the first and the second moments are time invariant. In other word, if the first moment is constant and if the covariance is not time dependent like show on equation 2 [27-29].

$$E[x_t] = \mu(t) = \mu \ and \ cov[x_t, x_{t+h}] = E[(x_t - \mu)(x_{t+h} - \mu)] = \gamma(h) \ \forall t, h \qquad (Eq\ 2)$$

Note that an equivalent stationarity criterion must be fond with the simple correlation coefficient (*corr*). The relation linking the two parameters is $cov[x_t, x_{t+h}] = corr[x_t, x_{t+h}] - \mu^2$. A stronger criterion is that the whole distribution (not only the mean and the variance) of the process does not depend on the time. The probability distribution $F$ of the stochastic process $x_t$ is invariant under a shift in time. In this case the series is called strict stationary [26], the two moments shown in the Equation 2 are stationary, but an other condition is also necessary (Equation 3).

$$F(x_1, ..., x_t,) = F(x_{1+h}, ..., x_{t+h}) \qquad (Eq\ 3)$$

The stationarity hypothesis is an important tool in classic time series analysis. As it is primordial for the ARMA method, this rule stays also correct for neural network studies [29]. In fact, all artificial networks are considered like functions approximation tools on a compact subset of $\mathbb{R}^n$. Moreover standard MLP (with at least 1 hidden layer) are asymptotically stationary, it converges to its stationary distribution, (i.e. $lim_n f_n = f$ ). Moreover, this kind of network can approximate any continuous and multivariate function. They cannot show "explosive" behavior or growing variance with time [21,30]. In practice a varying process may be considered to be close to stationary if it varies slowly and it is the modeling condition to use the MLP. Note that, the network can be trained to mimic a non-stationary process on a finite time interval. But the out-of-sample or prediction performance will be poor. Indeed, the network inherently cannot capture some important features of the process. Without pre-process, ANN and ARMA can be unappealing for many of the non-stationary problems encountered in practice [31]. One way to overcome this problem is to transform a non-stationary series into a stationary (weakly or stronger if possible) one and then model the remainder by a stationary process.

In our case, we have developed a sophisticated method to make the global radiation stationary ($X_t$, modeling of cycles). The original series has two periodicities (angular frequency $\omega_1$ and $\omega_2$) very difficult to overcome (Equation 4) because the amplitude $a$(t), $b$(t) and $R_t$ the cloud occurrence, have not simple expressions. Without the last term, this equation represents a clear sky estimation.

$$X_t = (a(t)\cos(\omega_1.t) + b(t)\cos(\omega_2.t)).R_t \qquad (Eq\ 4)$$

The first periodicity is a classic yearly seasonality which can be erased with a ratio to trend: multiplicative scheme induced by the nature of the global radiation series on the Equation 5. In previous studies [13], we have demonstrated that the clear sky index obtained with Solis model [5,32] is the more reliable for our locations. As the second daily seasonality is often not completely erased after this operation, then we use a method of seasonal correction (corrected for seasonal variance) based on the moving average [27,28]. The chosen method is essentially interesting for the case of a deterministic nature of the series seasonality (true for the global radiation series) but not for the stochastic seasonality [33]. The steps we follow to make the series stationary are as following:

   1-use of Solis model to establish the clear sky model of the considered location :

$$H_{gh}(t) = H'_0(t).exp\left(\frac{-\tau}{sin^b(h(t))}\right).sin(h(t)) \quad (Eq\ 5)$$

2-calculate the ratio to trend to overcome the periodicity, the result is the clear sky index CSI :

$$CSI_t = X_t/H_{gh}(t) \quad (Eq\ 6)$$

3-calculate the moving average (MM(t)) considering that $2.\Delta T + 1$ is equal to the periodicity of the series. In the case of a 9 hours periodicity, $\Delta T$ corresponds to 4 hours

$$MM_t = \langle CSI_t \rangle_{t\in[t-\Delta t, t+\Delta t]} \quad (Eq\ 7)$$

4-operate a new ratio to trend with the moving average. The new coefficient are called the periodic coefficients *C(t)* :

$$C_t = CSI_t/MM_t \quad (Eq\ 8)$$

5-Compute of the average over one year $E[C_t]$ (done on 365x9=3285 values because only 9 hours per day have been considered)

6-Compute of the new stationary series $CSI^*(t)$

$$CSI^*_t = CSI_t/E[C_t] \text{ (modulo 3285 hours)} \quad (Eq\ 9)$$

The Figure 2 shows the stationarity methodology concerning the global radiation time series prediction.

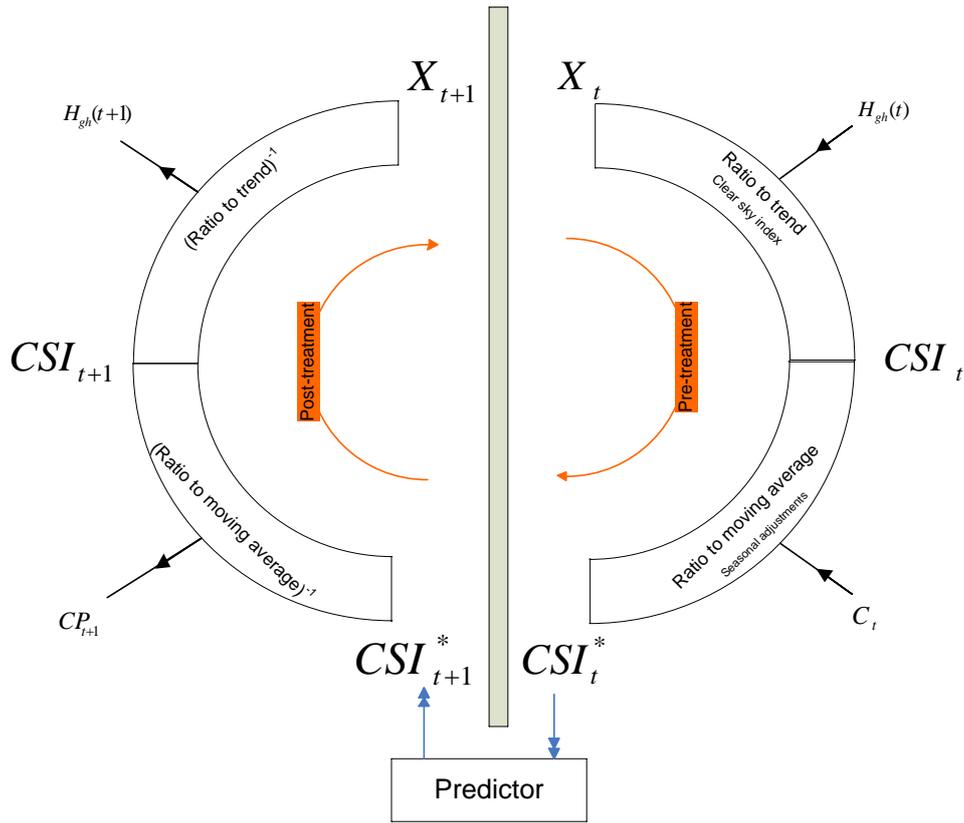

**Figure 2 : Scheme of the stationarity methodology**

To validate the stationary processing, we have chosen two tools. The first is the variation coefficient of the series ($VC_X = \sqrt{E[(x_t - \mu)^2]}/E[x_t] = \sigma/\mu$). Although this tool is easy to use, it is only dedicated to the cross comparison. It is not an absolute criterion allowing to distinguish the stationary and non-stationary process. To overcome this problem, there are a lot of available non-stationary tests: variance test of Fisher, unit root, Dickey-Fuller, KPSS, just to name a few. For more details, the reader can refer to Hamilton, 1994, Pollock, 1999 and Bourbounais, 2008 [27,29,31]. The quality of each method seems equivalent and some tests are not adapted to the seasonal case but to the extra-seasonality trends or differency stationary. We have chosen a very classic statistic test based on variance analysis: the Fisher test detailed by Bourbounais in 2008 (variance ratio to period by residue). This test involves a TS without trend as it is the case for the insolation. As matter of fact its annual average value is relatively constant at long term. In case of the periodic effect is significant, we can consider that the TS has a seasonal component. This test is proposed for the seasonality of the TS: daily and yearly phenomena. Before to perform this test it is necessary to construct a new scheme of the series, so $CSI_{i,j}^* = m_{i,j} + \epsilon_{i,j}$, with $m_{i,j} = a_{i,j} + b_{i,j}$ (*a* and *b* are respectively related to the daily and yearly seasonality), where $\epsilon_{i,j}$ is the associated residue described by a normal law $\rightarrow N(0, \sigma^2)$. The test is constructed from the null hypothesis H0 meaning no periodic influence in the TS, and the alternative hypothesis $H_1$ meaning TS with periodic influence. While details can be found in [27], principle of the Fisher test is described by the following equations:

1-Compute of the "empirical" Fisher statistic $F_c = V_p/V_R$ ($V_p$ the variance on period on the Equation 10 and $V_R$ the residue variance on the Equation 11). *p* is the number of measures per period and *N* the number of periods.

$$V_p = \frac{1}{p-1} \cdot \sum_{j=1}^{p} N \cdot (\langle CSI_{i,j}^* \rangle_i - \langle CSI_{i,j}^* \rangle_{i,j})^2 \qquad (Eq\ 10)$$

$$V_R = \frac{1}{(p-1)(N-1)} \cdot \sum_{i=1}^{N} \sum_{j=1}^{p} (CSI_{i,j}^* - \langle CSI_{i,j}^* \rangle_j - \langle CSI_{i,j}^* \rangle_i + \langle CSI_{i,j}^* \rangle_{i,j})^2 \qquad (Eq\ 11)$$

$$\left. \begin{array}{l} \langle CSI_{i,j}^* \rangle_j = \left(\frac{1}{p}\right) \sum_{j=1}^{p} CSI_{i,j}^* \ ; \\[6pt] \langle CSI_{i,j}^* \rangle_i = \left(\frac{1}{N}\right) \sum_{i=1}^{N} CSI_{i,j}^* \ ; \\[6pt] \langle CSI_{i,j}^* \rangle_{i,j} = \left(\frac{1}{Np}\right) \sum_{i=1}^{N} \sum_{j=1}^{p} CSI_{i,j}^* \end{array} \right\} \qquad (Eq\ 12)$$

For the *a* time series (daily effect) the parameter *N*=6 and *p*=3285, for the *b* series (yearly effect), *N*=6*365=2190 and *p*=9 (not 24 hours because only the hour between 8:00AM and 04:00PM are considered).

2-determination of the critical Fisher value for the degree of freedom $v_1 = (p-1)$ et $v_2 = (N-1).(p-1)$ and the $\alpha = 0,05$, so the value corresponding to:

$$F_{v_1;v_2}^{0,05} = F_{limit} \qquad (Eq\ 13)$$

3-If $F_c > F_{limit}$ the H$_0$ hypothesis is rejected and H$_1$ is accepted. If The $F_c$ calculated from the data is greater than the critical value of F distribution for desired false-rejection probability ($\alpha$= 0.05), the TS is described as seasonal. Furthermore, we can estimate than, more the $F_c$ coefficient is important, and more the seasonality component is important.

In section dedicated to the experiments we will present the results obtained for the variation coefficient and the Fisher test on our CSI stationary processing with and without periodic coefficients for the five places studied.

### b. ARMA

The ARMA method is certainly the most used with the prediction problems [29,34,27]. The ARIMA techniques are especially reference estimators in the prediction of global radiation field. It is a stochastic process coupling autoregressive component (AR) to a moving average component (MA). This kind of model is commonly called ARMA (p, q) and is defined with p and q parameters (Equation 14).

$$\left(1 - \sum_{t=1}^{p} \varphi_i . L^i\right). x_t = \left(1 + \sum_{t=1}^{q} \theta_i . L^i\right). \epsilon(t) \qquad (Eq\ 14)$$

Where, $x_t$ is a time series, $\varphi$ and $\theta$ are the parameters of the autoregressive and moving average part, L is the lag operator and ε is an error term distributed as a Gaussian white noise. The optimization of these parameters must be made depending on the type of the series studied. In the presented study, we chose to use Matlab© software and the Yule-Yalker fitting method [29]. The criterion adopted to consider when an ARMA model 'fits' to the global radiation time series is the normalized root mean square error described by the Equation 15 [35].

$$nRMSE = \sqrt{E\left[(X - \hat{X})^2\right]} / \sqrt{E[X^2]} \qquad (Eq\ 15)$$

The prediction error is generated by the prediction of two years of radiation not used during the ARMA parameters calculation step. Several experiments are needed to obtain the best model. Residual auto-correlogram tests have been computed to verify that the error term is a white noise. Before to use this method of forecasting, the global radiation time series is made stationary with the method described in the section 3-a (clear sky index with seasonal adjustment) and then, centered and reduced.

The models after optimization are very simple, the ARMA(1,0) for Ajaccio, Bastia, Montpellier and Nice or ARMA(2,0) for Marseille are successful. In the case of the ARMA(2,0) the prediction can be expressed by the Equation 16.

$$\hat{X}(t+1) = \sum_{i=0}^{1} \varphi_i X(t-i) + \epsilon(t+1) \qquad (Eq\ 16)$$

### c. Neural network and time series forecasting

Although a large range of different architecture of ANNs is available [36,37], MultiLayer Perceptron (MLP) remains the most popular [38,39]. In particular, feed-forward MLP networks with two layers (one hidden layer and one output layer) are often used for modeling and forecasting time series. Several studies [33,40,41] have validated this approach based on ANN for the non-linear modeling of time series. To forecast the time series, a fixed number p of past values are set as inputs of the MLP, the output is the prediction of the future value [42,43]. Considering the initial time series equation (Equation 1), we can transform this formula to the non-linear case of one hidden layer MPL with *b* related to the biases, *f* and *g* to the activation function of the output and hidden layer, and $\omega$ to the weights. The number of hidden nodes (*H*) and the number of the input node (*In*) allow to detail this transformation. (Equation 17):

$$\widehat{CSI}_{t+1} = f(\sum_{i=1}^{H} y_i \omega_i^2 + b^2) \qquad (Eq\ 17)$$

$$y_i = g(\sum_{j=1}^{In} CSI_{t-j+1} \omega_{ij}^1 + b_i^1)$$

In the presented study, the MLP has been computed with the Matlab© software and its Neural Network toolbox. The characteristics chosen and related to previous work are the following: one hidden layer, the activation functions are the continuously and differentiable hyperbolic tangent (hidden) and linear (output), the Levenberg-Marquardt learning algorithm with a max fail parameter before stopping training equal to 5. This algorithm is an approximation to the Newton's method [40] and is represented by the Equation 18 ($J(\omega)$ is the jacobian matrix, $J^T(\omega)$ this transposed and $e(\omega)$ the error between the *N* simulations and the *N* measures).

$$\Delta\omega = (\omega)^{k+1} - (\omega)^k = \pm[J^T(\omega)^k J(\omega)^k + \mu^k I]^{-1} J^T(\omega)^k e(\omega)^k$$

$$J(\omega) = \begin{pmatrix} \frac{\partial e_1(\omega)}{\partial \omega_{11}^1} & \cdots & \frac{\partial e_1(\omega)}{\partial \omega_H^2} \\ \vdots & \ddots & \vdots \\ \frac{\partial e_N(\omega)}{\partial \omega_{11}^1} & \cdots & \frac{\partial e_N(\omega)}{\partial \omega_H^2} \end{pmatrix} \qquad (Eq\ 18)$$

$$e(\omega) = \begin{pmatrix} e_1(\omega) \\ \vdots \\ e_N(\omega) \end{pmatrix} = \begin{pmatrix} CSI_1 - \widehat{CSI}_1 \\ \vdots \\ CSI_N - \widehat{CSI}_N \end{pmatrix}$$

In our case the parameter m takes the value 0.1 and 0.001 when the error, respectively, decreases or increases. Inputs are normalized on {-0.9,0.9} because the truncated interval gives better results than the full interval for the conditions of this study. Training, validation and testing data sets were respectively set to 80%, 20% and 0% (Matlab parameters). These phases concern the four first years and the global solar radiation test forecasting the two last years. The prediction methodology used in this paper is presented on the Figure 3. The pre-input layer consists on endogenous data of radiation and meteorological forecasts (exogenous) from the ALADIN numerical weather prediction model. We can see the possibility to select only the interesting data in the input layer among endogenous data of radiation and exogenous data from ALADIN model.

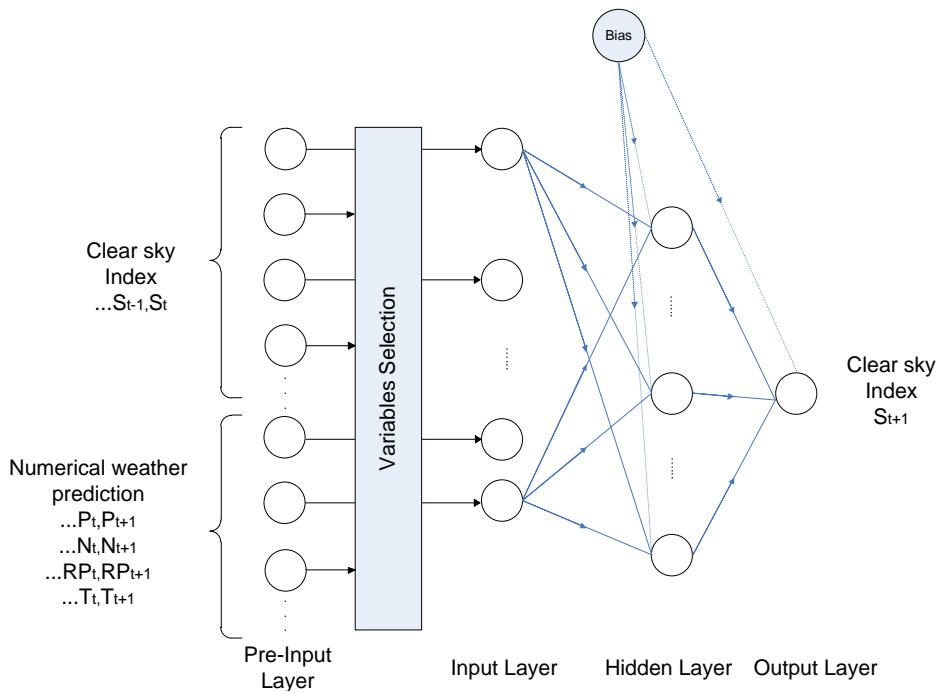

**Figure 3 : Scheme of the prediction methodology based on MLP model and NWP model**

The input variables selection step is one of the key tasks of ANN optimization. We proposed to base the input selection on the use of a regression model. The methodology used by our team in the past was to compute all the coefficients of correlation between exogenous data at time *t, t-1, etc.* and the clear sky index at time *t+1*. We encounter problems related to the difficulty to find a significant limit to this coefficient. To work around the apparent classical student T-test permissiveness and to respect the parsimony principle, we fixed empirically a limit not really justified theoretically. In the presented paper, we proposed a pre-input layer selection method in order to choose from a pool of available data. We have used the parsimony principle and limited the approach to 10 lags for the endogenous case and 2 lags for the each ALADIN output parameters. These 18 data in total are grouped in the pre-input Layer and are shown on the scheme of the Figure 2. This method also uses the statistical Student T-test to guide the choice but this time it is applied on the coefficients of a multiple linear regression. Note that concerning the ALADIN data, we consider that the physical signification of the lags upper than one are not very relevant because they are predictions of anterior time steps and not measures. In first we generate a regression model for the 18 pre-inputs nodes. There are 10 endogenous nodes (*In = 10*), and 2 for each meteorological variable called the total number of exogenous nodes is 8). The Equation 19 shows the simplest case of 10 endogenous nodes and *Me* nodes of exogenous parameter *E* [25]:

$$CSI^*(t+1) = \widehat{CSI^*}(t+1) + \epsilon(t+1)$$

$$= \sum_{j=1}^{In} \omega_j^{LR} X_{t-j+1} + \sum_{p=1}^{Me} \omega_{In+p}^{LR} E_{t-p+1} + b^{RL} + \epsilon_{t+1} \qquad (Eq\ 19)$$

This equation can be expressed by matrix form like shown in the Equation 20 or in the developed form in the Equation 21. $Y$ is a vector with the elements $CSI^*(t+1)$, $S'$ a matrix with concatenated endogenous and exogenous data and $W^{LR}$ a vector with all the adjustable parameters of the regression. For the estimation of this model the is $N_T$ data in the measurement history.

$$\begin{pmatrix} CSI^*_{t+1} \\ \vdots \\ CSI^*_{t+2-N_T} \end{pmatrix} = \qquad (Eq\ 20)$$

$$\begin{bmatrix} 1 & CSI^*_t & \cdots & CSI^*_{t-In+1} & E_t & \cdots & E_{t-Me+1} \\ 1 & \vdots & \ddots & \vdots & \vdots & \ddots & \vdots \\ 1 & CSI^*_{t-N_T+1} & \cdots & CSI^*_{t-In-N_T+2} & E_{t-N_T+1} & \cdots & E_{t-Me+N_T+2} \end{bmatrix} \cdot \begin{pmatrix} b^{RL} \\ \omega_1^{RL} \\ \vdots \\ \omega_{In+Me}^{RL} \end{pmatrix}$$

$$+ \begin{pmatrix} \epsilon_{t+1} \\ \vdots \\ \epsilon_{t+2-N_T} \end{pmatrix}$$

$$Y = S'W^{LR} + \epsilon \qquad (Eq\ 21)$$

The optimization corresponds to solve this equation to find correct values of linear regression weights. The least square method is used. The classical estimator is defined by $\widehat{W}^{LR} = (S'^T S')^{-1} S'^T Y$ ($S'^T$ is the transpose of the matrix $S'$). The dimension of the vector $W^{LR}$ is 19 (10 variables for the insolation, 8 for exogenous data, and 1 for the regression constant). The next step is to verify if some weights (related to each variable C$SI^*$, $N$, $RP$, $P$ and $T$ called $X_i$ for $1<i<12$) are not significantly different from zero. We use a Student T-test for all coefficients. The statistic used ($t_j$) corresponds to $t_j = \omega_j^{LR}/\sigma_j$ ($\sigma_j$ is the standard deviation of the parameter $j$). It is often easier to interpret it with the notation $t_j = \omega_j^{LR}/\sqrt{\epsilon^2(1/\sigma_j^2)}$ ($\epsilon$ is the error of prediction). The weight is not statistically equal to zero if $t_i$ is higher than the value 1.96 (large sample with 5% alpha level). Note that this test can be replaced by an equivalent one, based on the confidence interval of the weight considered ($\omega_j^{LR} \in [\omega_j^{LR} - t_j.\sigma_j, \omega_j^{LR} + t_j.\sigma_j]$). If the sign of the two limits are different, then we can assimilate the weight $\omega_j$ to zero. This is equivalent to search the sign of the produce of the two limits defined by $sign((\omega_j^{LR} + t_j.\sigma_j).(\omega_j^{LR} - t_j.\sigma_j)) = sign((\omega_j^{LR})^2 - t_j^2\sigma_j^2))$. The value must be strictly positive in order to consider $\omega_j$ different to zero. The variables with an associate weight equal to zero are the lest correlated with the insolation at the lag $t+1$. It is the criterion chosen to select (or not) the variable in the input layer of the neural network. In fact our pre-input layer selection method can be resumed by the following rule: $if\ \omega_j^{LR} = 0\ then\ \omega_{i,j}^1 = 0\ \forall i$. The application of this rule allows reducing the dimension of the input layer.

After this step, it is necessary to optimize the number of hidden nodes. The technique used is relatively standard; it consists to try several configurations by varying the number of nodes. In previous experiments [13], we have seen that often the number of hidden nodes must be comparable to the number of input nodes. After optimization (choice of input and hidden nodes, activation function, etc.) the output of the network can be expressed by the following expression (Equation 22) with $\widehat{P}, \widehat{N}, \widehat{RP}\ and\ \widehat{T}$ and by the number of prediction of each meteorological data use (respectively $p$, $n$, $rp$ and $t$). These predictions done with the model ALADIN depend on latitude, longitude, orography, temperature, humidity, etc. [18].

$$\widehat{CSI}^*(t+1) = b^2 + \sum_{i=1}^{H \in [1-20]} \omega_i^2 \cdot tanh\left(b_j^1 \sum_{j=0}^{9} \omega_{i,j}^1 CSI^*(t-j) + \sum_{p=1}^{p \in [1-2]} \omega_{i,9+p}^1 \widehat{P}(t-p+2) + \right.$$

$$\sum_{u=1}^{n \in [1-2]} \omega_{i,9+ps+u}^1 \widehat{N}(t-u+2) +$$

$$\left. \sum_{v=1}^{rp \in [1-2]} \omega_{i,9+ps+n+v}^1 \widehat{RP}(t-v+2) + \sum_{w=1}^{t \in [1-2]} \omega_{i,9+ps+n+rp+w}^1 \widehat{T}(t-w+2)\right) \quad \text{(Eq 22)}$$

Note that some elements of the Matrix $W^1$ ($\omega_{i,j}^1$) are equal to zero after the methodology related to the linear regression presented previously. The computing time is about 5 minutes with a Core i5 3.20GHz and 8Go of RAM. This computing time includes the clear sky generation, the seasonal adjustment, the statistical test, the ANN training and the average prediction computed for five ANNs with different initializing.

### d. Hybrid methodology

As in primary studies we have demonstrated that performances of ARMA and MLP were similar (in univariate case), we decide to investigate the possible contribution of a hybrid methodology combining Solis model and seasonal adjustments, ARMA and ANN models with ALADIN forecast data. The idea of hybrid methodology was born from the observation of some cases, where the model ARMA, the model ANN and in generally the single predictors are not really efficient. It will be the case if an important accuracy is required or when the priorization (during a benchmark) of predictors is impossible. The solution to combining methods to understand and model the dynamic of the signal is then considered: it's the base of the data mining technics. Among the predictors hybridization, we can mention the ARMA-MLP [44,45], the fuzzy inferences-MLP [22], the wavelet-MLP [46], the Markov-MLP [7] or Bayes-MLP [47]. Our choice is oriented to the ARMA-PMC (multivariate) method with time series made stationary. Concerning this model, we have used the classical approach shown previously but the predictor is now formed by two sub-predictors. There is a multitude of possible arrangement to construct it. We have opted to a model selection based on the transition linear/non-linear of the global radiation process. Linearity test exists (Lagrange multipliers), but are not really efficient when the two phenomena are concomitant in the same series. Our hypothesis is very simple, it consists to consider that when the nebulosity is low, the series is of linear character and the days with an important cloud occurrence, the series is of non-linear character. Thus, for the linear approach we use ARMA and for the non-linear approach, we select the PMC estimation. The validation of the hypothesis is done in an in-press article not yet published.. For Zhang [45] a hybrid model having both linear and nonlinear modeling abilities could be a good alternative for predicting time series data. By combining different models, different aspects of the underlying patterns may be captured. To construct the model selection we could only considered the day position in the year: during the winter months PMC would be selected and ARMA during the summer months. In this case, we would be considered a selection by rigid seasonality. This approach will not consider the sunny day in winter or the cloudy days in summer (although they are few). Consequently, we decided to design a methodology based on a not rigid, not repetitive and not well-marked seasonality. The new model is constructed as a stochastic model, which depends solely on the mistake made the previous hour. If the ARMA method was better at time *t*, (equivalent to sunny period, and indirectly to a linear phenomenon) then it will be still ARMA at *t+1* else it will be the MLP. We can summarize this method with the following rule (Equation 23) where $\varepsilon$ is the residue of the prediction:

$$if\ |\varepsilon^{AR}(t)|\ \leq\ |\varepsilon^{ANN}(t)|\ then\ \hat{x}(t+1) = \hat{x}^{AR}(t+1)\ else\ \hat{x}(t+1) = \hat{x}^{ANN}(t+1)\ (Eq\ 23)$$

The Figure 4 details the process followed by the hybrid methodology.

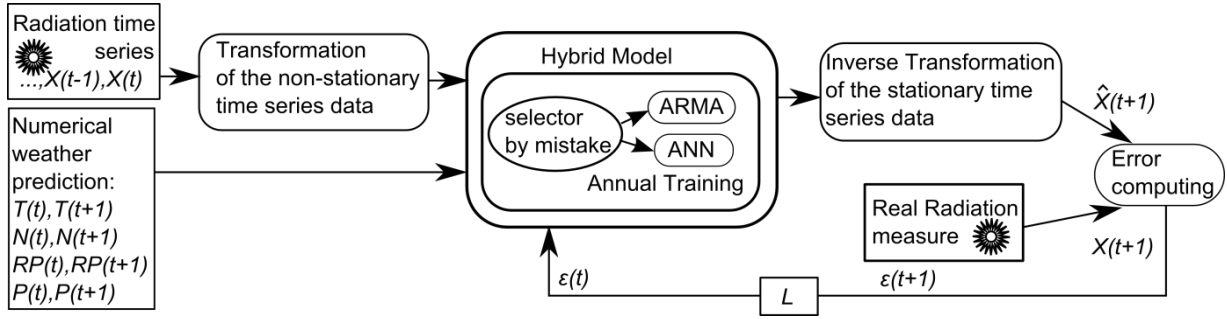

**Figure 4: Scheme of the hybrid method forecasting (the L-box is the lag or backshift operator)**

The next section presents, results obtained using this hybrid model on the five places located in Mediterranean area presented in section 2.

## 4. Experiments, results and discussion

This section includes all the experiments and results conducted during this study. The first part of this section is dedicated to the stationary efficiency of our preprocessing methodology. The second part shows intermediate results on MLP trained with endogenous and ALADIN forecast data. Finally we presents results obtained with the hybrid model presented in the previous section.

### a. Experiences about series makes Stationary

As we have seen previously in section 3.a, the different methods dedicated to the time series forecasting have to be used with stationary time series. It should be emphasized that if the available data are nonstationary, they must be made stationary before applied to ARMA or ANN model. Based on this observation, it is necessary to study stationary methods and to see if it should be interesting in the case of the global radiation time series forecasting. The two tested processes, the clear sky index (CSI) and the CSI with use of seasonal correction and periodic coefficients (CSI$^*$) are compared to the original series without treatment. To know if the preprocessing makes the series stationary, we use criterions like the variation coefficient (VC$_X$) considered as a dispersion rate and the Fisher T-test related to the daily and yearly stationarity. The values given on the Table 1 are established after a normalization process between {-0.9,0.9}. According to the Fisher Table, we found that threshold of the F-Test is $F_{\infty,\infty}^{0.05} = 1$ in the

|  |  | **Ajaccio** | **Bastia** | **Montpellier** | **Marseille** | **Nice** |
|---|---|---|---|---|---|---|
| **Original series** | VC | -3.45 | -2.48 | -3.36 | -3.62 | -3.21 |
|  | $F_{c(yearly)}$ | 12.99 | 12.44 | 11.12 | 15.07 | 13.72 |
|  | $F_{c(daily)}$ | 6.81 | **1.01** | 7.02 | 6.53 | 7.27 |
| **CSI** | VC | -0.40 | -0.4 | -0.37 | -0.38 | -0.42 |
|  | $F_{c(yearly)}$ | 3.25 | 3.05 | 3.39 | 4.08 | 3.44 |
|  | $F_{c(daily)}$ | 2.06 | 2.07 | 2.07 | 2.07 | 2.08 |
| **CSI*** | VC | -0.09 | -0.08 | -1.42 | -0.76 | -1.2 |
|  | $F_{c(yearly)}$ | **0.83** | **0.87** | **0.72** | **0.70** | **0.81** |
|  | $F_{c(daily)}$ | 2.05 | 2.04 | **1.41** | **1.80** | **1.52** |

Table 1: Stationary efficiency of CSI and CSI* preprocessing for the five cities studied. Bold values signify that the stationarity is effective.

The $VC_X$ parameter is not the more interesting as a stationarity criterion but it is very simple to use. The two other parameters show that the CSI* processing is the most efficient. There is only one case where the $F_c$ is not minimal with the CSI* (Bastia, $F_{c(daily)}$=2.04). Indeed, with this method, 8 of 10 parameters (two parameters by city) are lower than the Fisher thresholds. Note that with the others cases, only one empirical Fisher parameter indicates a quasi-stationarity in Bastia ($F_{c(daily)}$=1.01). It is certainly because on this location the weather is very unstable, generating a very noisy global radiation time series. On the Figure 5, we can see the impact of the CSI and CSI* preprocessing on a global radiation time series for the difficult Bastia case study. We obtained similar results for the four other locations.

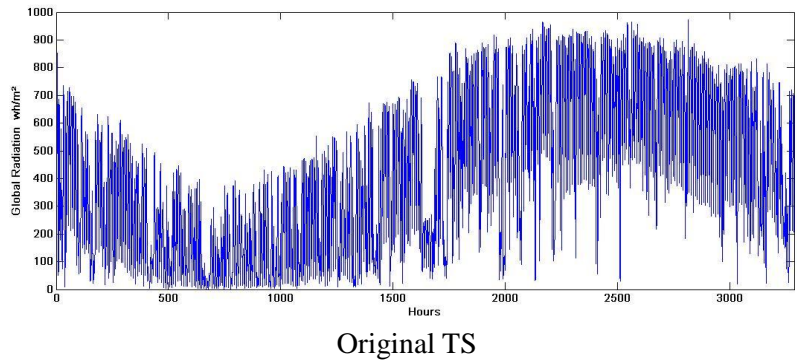

Original TS

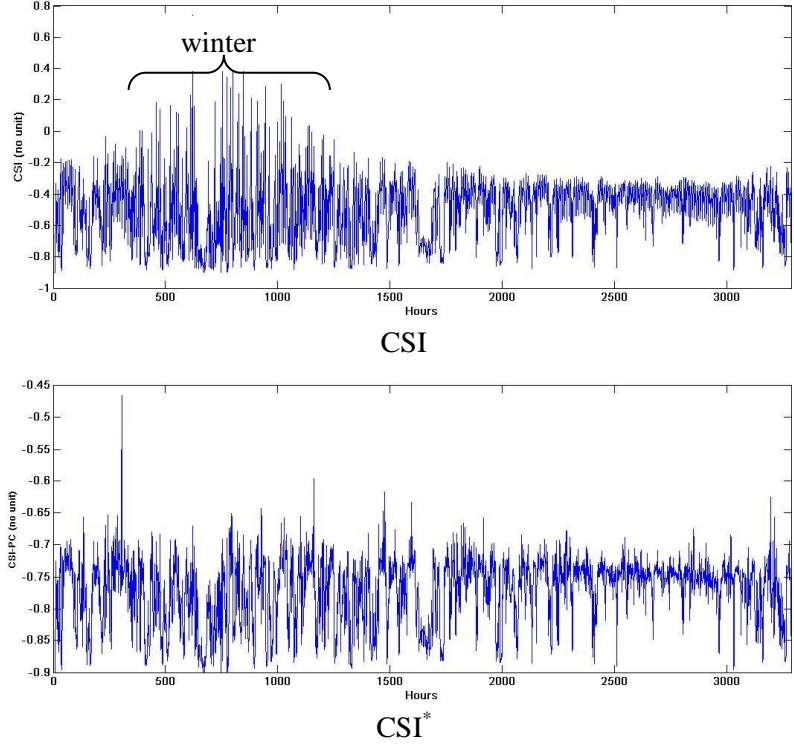

**Figure 5**: Effects of the stationarity processing on the Bastia time series during 365 days (oct 2002-oct 2003). The CSI* and the CSI are normalized.

While the CSI* processing seems the most interesting, the CSI seems graphically give the first moment constant (average fixed around the year ~-0.4), but let the second central moment about the mean of the series with periodicity (variance more important during winter than summer). We can see that the curve related to the CSI* process seems the most non-seasonal, contrary to the CSI process. In winter the standard deviation has not increased. The results presented in the next section will argue if the stationarity of the series increases the prediction quality.

### b. Results with MLP and ALADIN forecast data

This subpart proposes to analyze the impact of adding data from the ALADIN numerical model to a MLP. We have chosen to compare it with four other models (see Table 2). The first model which is considered as the reference is based on ARMA, the second is an endogenous MLP with our clear sky index preprocessing (CSI), the third is an endogenous MLP with CSI and seasonal correction (CSI*), and the last is an MLP with ALADIN forecast data. In the following, we adopt a canonical form to present the MLP architecture obtained after optimization steps: $(endo^e, RP^{rp}\ N^n, P^p, T^t) \times H \times S$, where $e$,

*rp, n, p* and *t* are the numbers of neurons activated in relation with endogenous data, precipitation, nebulosity, pressure and temperature, *H* and *S* are the number of hidden and output nodes. The exact description of the compared methods is:

**I.** ARMA+PC : The best ARMA model with the CSI* preprocessing;

**II.** ANN : The best endogenous MLP with the CSI preprocessing. Optimizations have been done with a standard method based on the interpretation of PACF partial autocorrelation factor [17];

**III.** ANN+PC: The best MLP with CSI* preprocessing. Optimizations have been done with a standard method based on the autocorrelation factor and cross-correlation;

**IV.** ANN+ALADIN: An optimized MLP with the CSI preprocessing and pre-input layer selection method using ALADIN forecast data in input;

**V.** ANN+ALADIN+PC: Same as previous but with CSI* preprocessing.

|   | Models | Annual | Winter | Spring | Summer | Autumn |
|---|---|---|---|---|---|---|
| **Ajaccio** | Persistence | 25.1 | 34.7 | 25.2 | 21.4 | 33.9 |
|   | **I. ARMA(1,0)** | 19.4 | 29.4 | **17.7** | 14.3 | 26.8 |
|   | **II. ANN** (Endo$^{1-10}$)x15x1 | 20.3 | 27.2 | 20.4 | 13.7 | 24.1 |
|   | **III. ANN+ PC** (Endo$^{1,2,3,4}$)x15x1 | 18.6 | 25.3 | 18.4 | 12.2 | 24.2 |
|   | **IV. ANN +ALADIN** (Endo$^{1,2,5,6,10}$PR$^{1,2}$N$^{1,2}$T$^{1,2}$)x15x1 | 19.0 | 26.8 | 19.1 | 12.3 | 23.0 |
|   | **V. ANN +ALADIN+PC** (Endo$^{1,5,9,10}$PR$^{1,2}$N$^{1}$P$^{1,2}$T$^{1,2}$) x15x1 | **17.8** | **24.9** | 18.1 | **11.7** | **22.0** |
| **Bastia** | Persistence | 27.1 | 35.0 | 27.1 | 22.6 | 34.4 |
|   | **I. ARMA(1,0)** | 21.1 | 26.7 | **20.3** | 15.8 | 26.9 |
|   | **II. ANN** (Endo$^{1-10}$)x10x1 | 22.8 | 27.3 | 23.4 | 16.1 | 25.7 |
|   | **III. ANN+PC** (Endo$^{1,2,3,4}$)x15x1 | 20.8 | 24.9 | 21.4 | 14.9 | 24.9 |
|   | **IV. ANN +ALADIN** (Endo$^{1,3,8:10}$PR$^{1}$N$^{1,2}$P$^{1,2}$T$^{1,2}$) | 21.3 | 25.8 | 21.7 | 15.1 | 24.0 |
|   | **V. ANN +ALADIN+PC** (Endo$^{1,3,5}$PR$^{1,2}$N$^{1,2}$P$^{1,2}$T$^{1,2}$) | **19.9** | **24.4** | 20.5 | **14.2** | **23.3** |
|   | Persistence | 26.9 | 32.6 | 25.9 | 24.6 | 33.2 |
|   | **I. ARMA(1,0)** | 20.1 | 23.5 | 18.7 | **15.5** | 21.9 |
|   | **II. ANN** (Endo$^{1-10}$)x10x1 | 20.8 | 22.4 | 20.2 | 17.9 | 19.3 |
| | III. ANN+PC<br>(Endo$^{1,2,3}$)x15x1 | 19.3 | 20.4 | 18.8 | 16.0 | 19.8 |
| | IV. ANN +ALADIN<br>(Endo$^{1,3,6:10}$Pr$^{1,2}$N$^{1,2}$P$^{1,2}$T$^{1,2}$) | 19.3 | 20.3 | 18.6 | 16.8 | **18.1** |
| | V. ANN +ALADIN+PC<br>(Endo$^{1,5,10}$Pr$^{1,2}$N$^{1,2}$P$^{1,2}$T$^{1,2}$) | **18.6** | **20.1** | **17.9** | **15.5** | 19.2 |
| Marseille | Persistence | 25.3 | 32.9 | 25.3 | 20.0 | 32.3 |
| | I. ARMA(2,0) | 18.9 | 23.9 | 19.0 | 11.8 | 21.4 |
| | II. ANN<br>(Endo$^{1-10}$)x10x1 | 19.0 | 22.5 | 20.7 | 11.3 | 18.8 |
| | III. ANN+PC<br>(Endo$^{1,2,3,4}$)x15x1 | 16.9 | 20.6 | 17.8 | 10.5 | 17.1 |
| | IV. ANN +ALADIN<br>(Endo$^{1,2,6:10}$PR$^{1,2}$N$^{1,2}$P$^{1,2}$T$^{1}$) | 17.4 | 20.4 | 18.5 | 10.4 | **16.4** |
| | V. ANN +ALADIN+PC<br>(Endo$^{1,3,7}$PR$^{1,2}$N$^{1,2}$P$^{1,2}$T$^{1,2}$) | **16.3** | **19.6** | **16.6** | **10.3** | 16.4 |
| Nice | Persistence | 26.4 | 32.1 | 24.5 | 21.1 | 37.1 |
| | I. ARMA(1,0) | 20.7 | 23.5 | **17.6** | 12.4 | 37.5 |
| | II. ANN<br>(Endo$^{1-10}$)x10x1 | 20.9 | 21.7 | 18.8 | 11.7 | 32.3 |
| | III. ANN+PC<br>(Endo$^{1,2,3}$)x15x1 | 20.1 | 20.5 | 19.1 | 11.4 | 30.9 |
| | IV. ANN +ALADIN<br>(Endo$^{1,3,7:10}$PR$^{1}$N$^{1}$P$^{1,2}$T$^{1,2}$) | 20.1 | 20.4 | 18.1 | 11.4 | 32.3 |
| | V. ANN +ALADIN+PC<br>(Endo$^{1:5}$PR$^{1,2}$N$^{1,2}$P$^{1,2}$T$^{1,2}$) | **19.4** | **19.9** | 18.6 | **11.0** | 30.1 |

Table 2: Comparison between five forecasting methods for the five cities studied. The error is the nRMSE and best results are in bold.

If we analyze the annual error, the methodology V, based on the utilization of an optimized MLP with ALADIN forecasting data and seasonal adjustment with periodic coefficients (CSI$^*$ preprocessing) is the most relevant. If we analyze seasonally the errors, we find only four cases (Ajaccio, Bastia and Nice during spring and Montpellier during summer) which have a better forecaster: ARMA method. For the other cases, the ARMA method seems equivalent to the ANN method without pretreatment (method II) but is worse than the ANN when the stationary of the series is the same (method III). While the distinct adding of CSI$^*$ preprocessing (method III ; error is 13.0%) and ALADIN forecast data (method IV error is 13.2%) to the ANN model improves the prediction, their combined adding (method V) potentiates this effect. Because the sun light is abundant and there is low occurrence of clouds, the summer season is the most favorable to prediction.

### c. Results with the hybrid methodology

This subpart proposes to present results of the hybrid method (see the Table 3) defined in section 3.d. The ARMA and MLP models used are those presented in the previous section: models I and V. This methodology is based on the prediction of "ARMA" and "ANN" with ALADIN data, clear sky index and seasonal adjustments. The ratio ARMA/ANN represents the numbers of hourly ARMA simulations versus ANN simulations We can observe that the distribution of predictor use is 1/3 for ARMA and 2/3 for ANN. This observation confirms the results previously presented. In summer, the error is for the first time lower than 10% (Marseille). All the seasons and cities benefit of the use of this method coupling ARMA and ANN. The summer is the season where the gain is the less significant, certainly because it is the season where the stand-alone methodologies (ANN and ARMA)

|  | Ratio ARMA/ANN | Annual | Winter | Spring | Summer | Autumn |
|---|---|---|---|---|---|---|
| **Ajaccio** | 2592/3978 | 14.9 | 19.4 | 15.5 | 11.0 | 17.0 |
| **Bastia** | 2557/4013 | 16.5 | 19.5 | 17.5 | 13.2 | 17.9 |
| **Montpellier** | 2348/4222 | 14.7 | 15.7 | 15.2 | 13.4 | 15.5 |
| **Marseille** | 2124/4446 | 13.4 | 16.6 | 14.8 | 9.3 | 13.8 |
| **Nice** | 2301/4269 | 15.3 | 16.6 | 15.3 | 10.3 | 26.2 |

**Table 3 : Annual and seasonal results for the hybrid ARMA/ANN model**

Ultimately, We can see that all the methodologies proposed in this paper have decreased the prediction error. In average on the five cities, the total nRMSE decrease of 11.3% against a naïve persistence predictor which has a average nRMSE equal to 26.2%. A step very interesting is the mixing between the ANN and ARMA, but do not forget that this result is only true with the type of ANN and ARMA considered. In the Figure 6, we can see the matching between measure and simulation (hybrid method presented previously) for all the cities. We can presume that there is a strong correlation between the two quantities. Except to Bastia, the cloudy periods (like the 71°-81° hour interval) seem correctly predicted. Concerning the case of Bastia, the model is not able to anticipate the nebulosity, and improvements are certainly necessary.

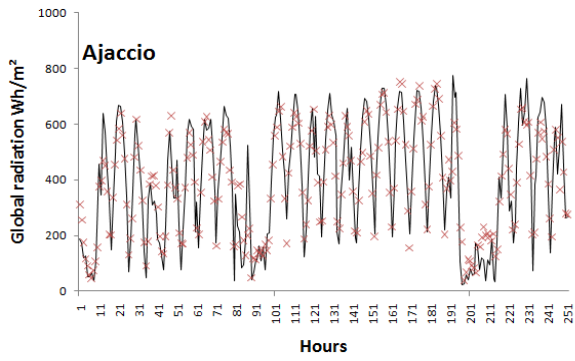
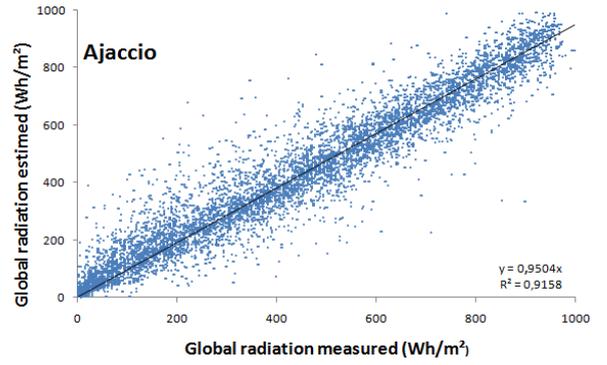

5.

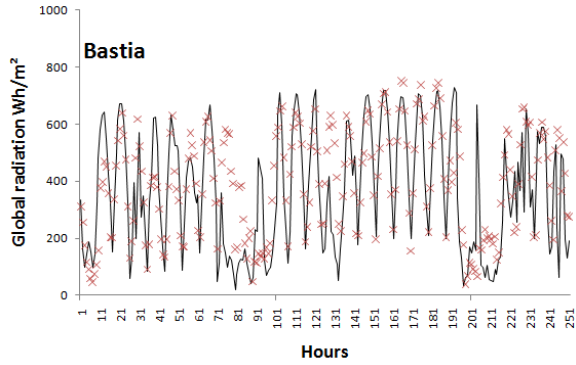
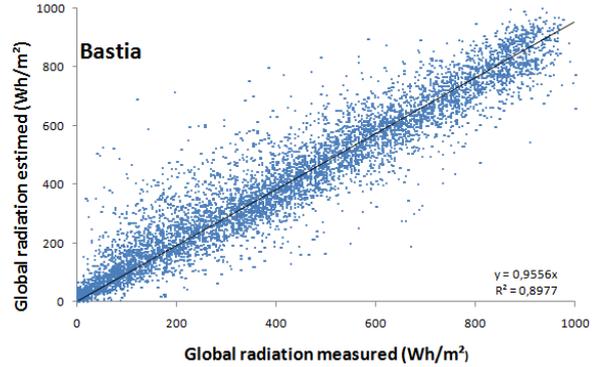

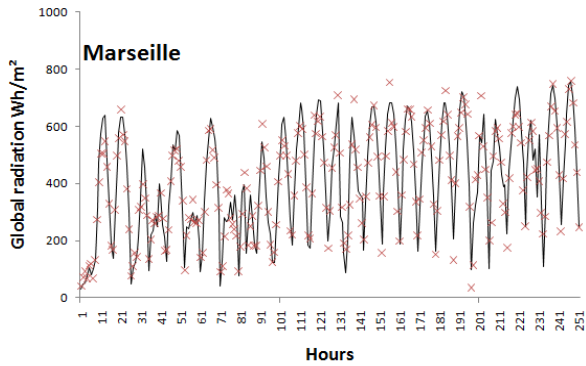
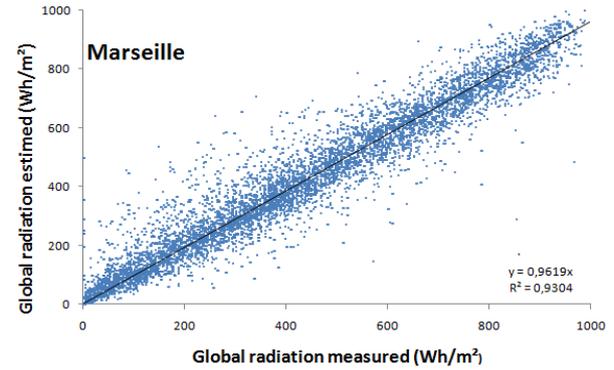

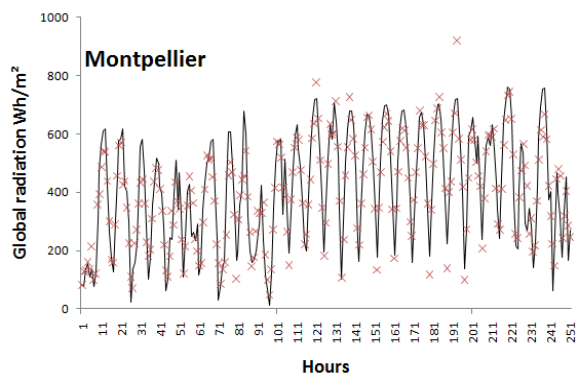
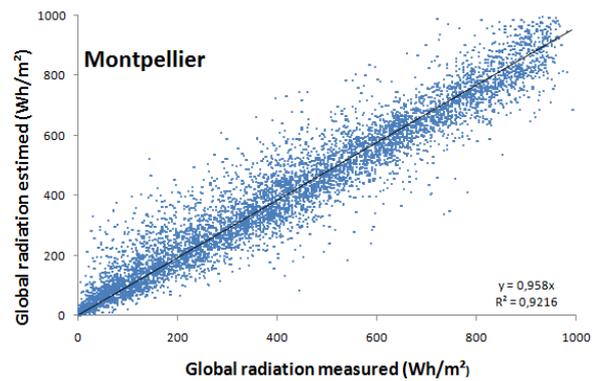

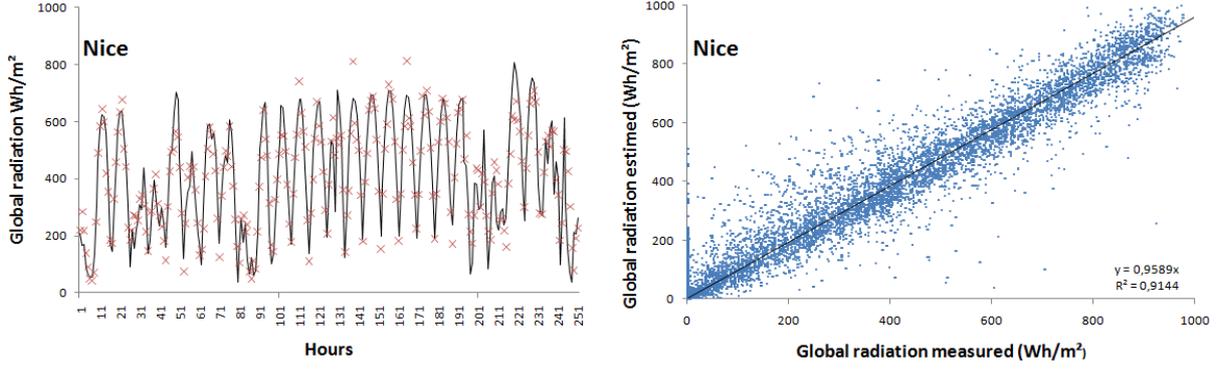

**Figure 6: Comparison between measured and simulated global radiation done with the hybrid methodology "ARMA" and "ANN+ALADIN+PC+CSI".The 250 hours shown corresponding to half-February to half-March 2008)**

### a. Evaluation of the prediction relevance

Considering that the prediction methodology proposed here is mainly designed for a power manager, it is necessary to couple the forecasted value to a confidence interval. The presented study proposes to compute it during the training step of the MLP, and then to use it during the prediction. The simulator gives for each hour two parameters: the h+1 horizon global radiation and a parameter representing the confidence we can give to this value. Before to compute this parameter it is necessary to explain the parameter *CI(t)* represented in the equation 24. In fact it is the absolute residue error of the prediction during the training sample. The training set includes 4 years, and each year includes 365x9=3285 hours, so 4x3285 elements. Then an hourly average is done allowing to transform this *CI(t)* series to a new series *CI\*(t)* like described in the equation 25.

$$CI(t) = |\epsilon(t)| \; \forall t \in [1, 4\times 3285] \quad (Eq\ 24)$$

$$CI^*(t) = \left(\frac{1}{4}\right).\sum_{i=1}^{4} CI(t + (i-1).3285)) \; \forall t \in [1; 3285] \quad (Eq\ 25)$$

This index is necessary to judge the relevance of the prediction. For example, the Figure 7 shows for Ajaccio the prediction and the confidence interval (average ± CI). Three models are evaluated: the ANN with only endogenous data, the ANN with ALADIN and the periodic coefficients and the hybrid model "ARMA" with "ANN+ ALADIN+PC". The period considered is one of the most complicated to forecast. It corresponds to the winter with a lot of cloudy days. It is interesting to see that the simple model endogenous ANN describes well the radiation, however, there are some outliers points (15$^{th}$ or 41$^{th}$ hour), with a confidence interval not significant. In the second curve related to the ANN+ALADIN+PC, the error is most regular, the atypical points seems non-existent. The third curve (both of ANN and ARMA) is visually the most interesting, despite that the confidence interval is yet incorrect for some points. In fact, with the hybrid method (ARMA+ANN), the ANN forecast is preponderant during the cloudy days; the ARMA predictions are used only during the sunny days. So the two last methods presented are relatively similar for the month considered.

6.

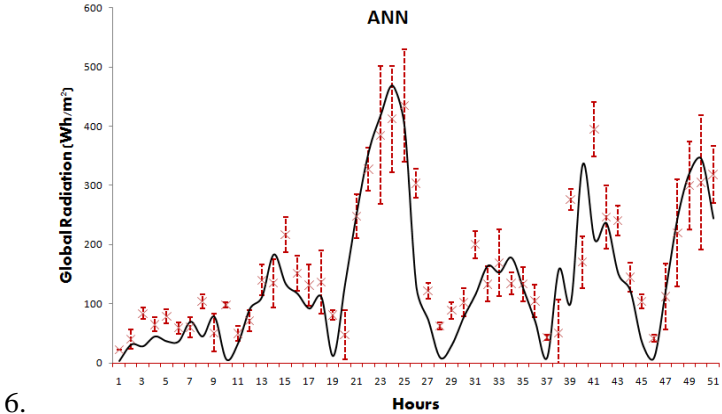

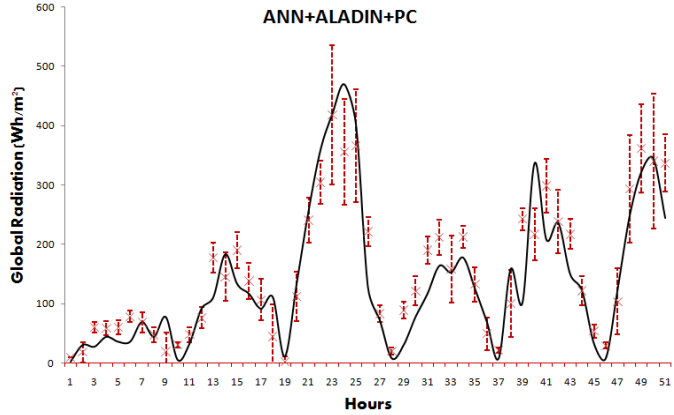

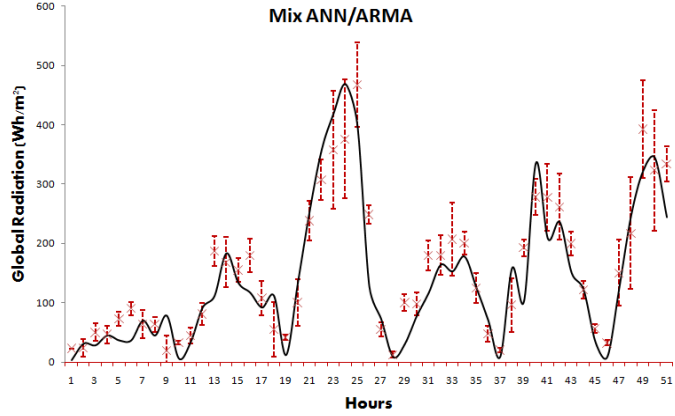

**Figure 7: Prediction marked with confidence interval of the global radiation in Ajaccio during the 50<sup>th</sup> first hours of January 2007. The line is related to the measure.**

## 7. Conclusion

We proposed in this paper an original technique to predict hourly global radiation time series using meteorological forecasts from a numerical weather prediction model. We optimized a Multi-Layer Perceptron (MLP) with ALADIN forecast data and endogenous data previously made stationary. We have used an innovative pre-input layer selection method and we have combined our optimized MLP to an Auto-Regressive and Moving Average (ARMA) model from a rule based on the analysis of hourly data series. This model has been successfully used to forecast the hourly global horizontal radiation for five places in Mediterranean area. In addition, this paper has allowed determining a stationarity process (method and control) for the global radiation time series. This step is primordial to correctly forecast the future values of insolation (annual average nRMSE gain over the five locations equals to 1.7%). The use of ALADIN forecasting data as an input of a MLP has shown a really great interest to improve the prediction (average nRMSE gain of 0.7%). These results could certainly be improved by a better comprehension of the complexity of the model and collaboration with a professional forecaster of Meteo-France who could help us in selection of data especially concerning the runs to consider. In last, the use of a hybrid method coupling ANN and ARMA predictors decrease much the prediction error (average nRMSE gain to 3.5%). If we compare the results with a standard model like persistence method (a typically naïve predictor) the nRMSE error is reduce by about 11.3%. The detail of the prediction error decrease following the different steps of the forecasting methodology is done on the Figure 8. The Last important point treated in this paper is the proposition of using confidence intervals in order to estimate the reliability of the prediction. The perceptive of this work are related the demonstration of the generalization of our model. We would show that the superior performance of this model is not likely to be a consequence of data mining (or data snooping). In fact, it should be sure that the model constructed in this way is not of limited practical value.

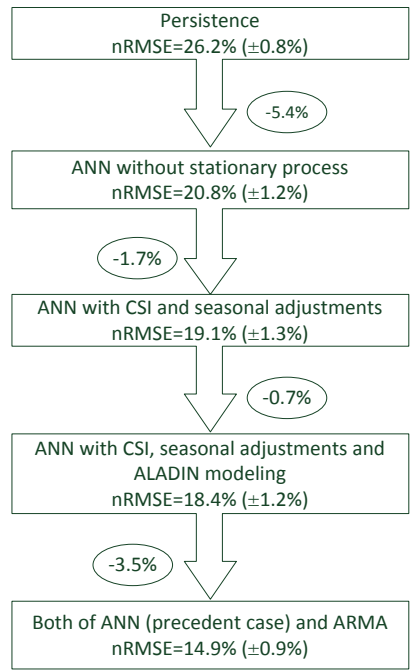

**Figure 8: Prediction error decrease following the different steps of the forecasting methodology**

Concerning the practical and policy implication of the results shown here, the conclusions are certainly compatible with the deployment prerogatives of renewable energy.

Because of the intermittent nature of some energy sources (like PV or wind energy), they are included in a limited way in power systems. This limitation ensures the electrical grid stability. There are two methodologies to overcome this limitation problem: the storage of the overflow to redistribute it adequately and the prediction of the energy sources. This last solution allows to use at the right time other energy productions not dependent on weather, so to avoid a drop of the electrical current or maybe a black-out. The main problem of the management is the dispatch between these two energy types. This study deals with the one hour forecasting horizon. It is certainly the most important horizon for an electrical manager because it corresponds to the starting delay of some conventional energy sources like diesel engines or gas turbines for power generation. Actually the industrially prediction of the global radiation is often done with a simple persistence. Even if, the predictor is effective in sunny days, the present study shows that more successful forecasters exist. This result is especially interesting in the insular case (or generally to overcome the isolation problem), where the interconnection is limited and where the energy autonomy must be considered at medium or long-term.

In future studies, it would be very relevant to study an approach for solar irradiance forecasting 24 hours ahead using several MLP connected, to decrease the time step and to test the methodology on a real PV module. In addition, an important work will be to simplify the model while keeping an acceptable prediction. Indeed, even if our method is attractive, it could be complex and costly to implement for an electric power manager in grid stability context of power supply mixing renewable and conventional energy.

# Acknowledgment

This work was partly supported by the Territorial Collectivity of Corsica. We thank the French National Meteorological Organization (Météo-France) and the CNRM (Centre National de Recherches Météorologiques) and particularly Ms. Delmas who has supervised the data collection from their data bank for the five synoptic stations and from the ALADIN model.

# List of captions

Figure 1. The five studied stations marked in the Mediterranean sea: Ajaccio, Bastia, Montpellier, Marseille and Nice.

Figure 2 : Scheme of the stationarity methodology

Figure 3 : Scheme of the prediction methodology based on MLP model and NWP model

Figure 4: Scheme of the hybrid method forecasting (the L-box is the lag or backshift operator)

Figure 5: Effects of the stationarity processing on the Bastia time series during 365 days (oct 2002-oct 2003). The $CSI^*$ and the CSI are normalized.

Figure 6: Comparison between measured and simulated global radiation done with the hybrid methodology "ARMA" and "ANN+ALADIN+PC+CSI".The 250 hours shown corresponding to half-February to half-March 2008)

Figure 7: Prediction marked with confidence interval of the global radiation in Ajaccio during the $50^{th}$ first hours of January 2007. The line is related to the measure.

Figure 8: Prediction error decrease following the different steps of the forecasting methodology

# List of tables

Table 1: Stationary efficiency of CSI and $CSI^*$ preprocessing for the five cities studied. Bold values signify that the stationarity is effective.

Table 2: Comparison between five forecasting methods for the five cities studied. The error is the nRMSE and best results are in bold.

Table 3 : Annual and seasonal results for the hybrid ARMA/ANN model

# Bibliographie